\documentclass{article}

\usepackage{microtype}
\usepackage{graphicx}
\usepackage{booktabs} 

\usepackage{helvet}  
\usepackage{courier}  
\usepackage{url}  
\usepackage{graphicx}  
\usepackage{multirow}
\usepackage{amsthm}
\usepackage{color}
\usepackage{MnSymbol}
\usepackage{makecell}
\usepackage{arydshln}
\usepackage{amsmath}
\usepackage{xcolor}
\usepackage{caption} 
\usepackage{natbib}
\usepackage{subcaption}
\usepackage{textcomp}
\usepackage{wrapfig}
\usepackage{algorithm}
\usepackage{algorithmic}
\usepackage{amsfonts}
\usepackage{mathtools}

\DeclareMathOperator{\E}{\mathbb{E}}

\newtheorem{theorem}{Theorem}[section]
\newtheorem{corollary}{Corollary}[theorem]

\newtheorem{definition}{Definition}[section]

\usepackage[accepted, nohyperref]{icml2020}

\AtBeginDocument{%
  \addtolength\abovedisplayskip{-0.3\baselineskip}%
  \addtolength\belowdisplayskip{-0.3\baselineskip}%
}

\icmltitlerunning{Learning Not to Learn in the Presence of Noisy Labels}

\begin{document}

\twocolumn[
\icmltitle{Learning Not to Learn in the Presence of Noisy Labels}



\icmlsetsymbol{equal}{*}

\begin{icmlauthorlist}
\icmlauthor{Liu Ziyin}{tk}
\icmlauthor{Blair Chen}{cmu}
\icmlauthor{Ru Wang}{tk}
\icmlauthor{Paul Pu Liang}{cmu}
\icmlauthor{Ruslan Salakhutdinov}{cmu}
\icmlauthor{Louis-Philippe Morency}{cmu}
\icmlauthor{Masahito Ueda}{tk,riken}
\end{icmlauthorlist}

\icmlaffiliation{tk}{Department of Physics \& Institute for Physics of Intelligence, University of Tokyo}
\icmlaffiliation{cmu}{Carnegie Mellon University}
\icmlaffiliation{riken}{RIKEN CEMS}

\icmlcorrespondingauthor{Liu Ziyin}{zliu@cat.phys.s.u-tokyo.ac.jp}

\icmlkeywords{Machine Learning, ICML}

\vskip 0.3in
]



\printAffiliationsAndNotice{}  

\begin{abstract}
Learning in the presence of label noise is a challenging yet important task: it is crucial to design models that are robust in the presence of mislabeled datasets. In this paper, we discover that a new class of loss functions called the gambler's loss provides strong robustness to label noise across various levels of corruption. We show that training with this loss function encourages the model to ``abstain" from learning on the data points with noisy labels, resulting in a simple and effective method to improve robustness and generalization. In addition, we propose two practical extensions of the method: 1) an analytical early stopping criterion to approximately stop training before the memorization of noisy labels, as well as 2) a heuristic for setting hyperparameters which do not require knowledge of the noise corruption rate. We demonstrate the effectiveness of our method by achieving strong results across three image and text classification tasks as compared to existing baselines.
\end{abstract}

\vspace{-6mm}
\section{Introduction}
\vspace{-1mm}


Learning representations from real-world data can greatly benefit from clean annotation labels. However, real-world data can often be mislabeled due to 1) annotator mistakes as a natural consequence of large-scale crowdsourcing procedures~\citep{Howe:2008:CWP:1481457}, 2) the difficulty in fine-grained labeling across a wide range of possible labels~\citep{russakovsky2015imagenet}, 3) subjective differences when annotating emotional content~\citep{journals/lre/BussoBLKMKCLN08}, and 4) the use of large-scale weak supervision~\citep{46574}. Learning in the presence of noisy labels is challenging since overparametrized neural networks are known to be able to memorize both clean and noisy labels even with strong regularization~\citep{Zhang_rethink}. Empirical results have shown that when the model memorizes noisy labels, its generalization performance on test data deteriorates (e.g., see Figure~\ref{fig:observation}). Therefore, learning in the presence of label noise is a challenging yet important task: it is crucial to design models that are robust in the presence of mislabeled datasets.

In this paper, we show that a new class of loss functions called the gambler's loss~\citep{ziyindeep} provides strong robustness to label noise across various levels of corruption. We start with a theoretical analysis of the learning dynamics of this loss function and demonstrate through extensive experiments that it is \textit{robust} to noisy labels. Our theory also motivates for two practical extensions of the method: 1) an analytical early stopping criterion designed to stop training before memorization of noisy labels, and 2) a training heuristic that relieves the need for hyperparameter tuning and works well without requiring knowledge of the noise corruption rate. Finally, we show that the proposed method achieves state-of-the-art results across three image (MNIST, CIFAR-10) and text (IMDB) classification tasks, compared to prior algorithms.  

\vspace{-3mm}
\section{Background and Related Work}
\label{sec: related work}
\vspace{-1.5mm}

In this section we review important background and prior work related to our paper.

\textbf{Label Noise:} Modern datasets often contain a lot of labeling errors~\citep{russakovsky2015imagenet,schroff2010harvesting}. Two common approaches to deal with noisy labels involve using a surrogate loss function~\citep{patrini2017making, zhang2018generalized, xu2019l_dmi} that is specific to the label noise problem at hand, or designing a special training scheme to alleviate the negative effect of learning from data points with wrong labels~\citep{yu2019does}. In this work, we mainly compare with the following two recent state-of-the-art methods: 
\textbf{Co-teaching+}~\citep{yu2019does}: this method simultaneously trains two networks which update each other with the other's predicted label to decouple the mistakes; \textbf{Generalized cross-entropy } ($\mathbf{L_q}$)~\cite{zhang2018generalized}: this method uses a loss function that incorporates the noise-robust properties of MAE (mean absolute error) while retaining the training advantages of CCE (categorical cross-entropy).

\textbf{Early Stopping}. Early stopping is an old problem in machine learning~\cite{prechelt1998early, Amari:1998:NGW:287476.287477}, but studying it in the context of label noise appeared only recently~\cite{li2019gradient, hu2019understanding}. It has been shown theoretically~\citep{li2019gradient} that early stopping can constitute an effective way to defend against label noise, but no concrete method or heuristic has been presented. In this paper, we propose an early stopping method that can be used jointly with the gambler's loss function. Our analysis on this loss function allows us to propose an analytic function to predict an early stopping threshold \textit{without using a validation set} and is independent of the model and the task, provided that the model has sufficient complexity to solve the task (e.g. overparametrized neural networks). 
To the best of our knowledge, we have proposed the first early stopping method effective for noise labels.

\textbf{Learning to Abstain}: Within the paradigm of \textit{selective classification}, a model aims to abstain from making predictions at test time in order to achieve higher prediction accuracy~\citep{el2010foundations}. Given a $m$-class prediction function $f:\mathcal{X}\to\mathcal{Y}$ and a selection function $g:\mathcal{X}\to \{0,1\}$, a selective classifier can be defined as
\begin{equation}
    (f,g)(x)\triangleq \begin{dcases}f(x), &\text{if }g(x)=1\\ \text{ABSTAIN}, &\text{if }g(x)=0.\end{dcases}
\end{equation}
Efforts to optimize such a classifier have evolved from a method to train $g$ given an existing trained $f$ to a multi-headed model architecture that jointly trains $f$ and $g$ given the desired fraction of data points~\citep{geifman2019selectivenet}. The gambler's loss represents a recent advancement in this area by jointly training $(f,g)$ by modifying the loss function and thus introducing a more versatile selective classifier that performs at the SOTA~\citep{ziyindeep}. 

\vspace{-1mm}
\subsection{Gambler's Loss Function}
\vspace{-1mm}

The gambler's loss draws on the analogy of a classification problem as a horse race with bet and reserve strategies. Reserving from making a bet in a gamble can be seen as a machine learning model abstaining from making a prediction when uncertain. This background on gambling strategies has been well studied and motivated in the information theory literature~\cite{CoverInformationTheory, UniversalPortfolio}, and was recently connected to prediction in the machine learning literature~\cite{ziyindeep}.

We provide a short review but defer the reader to~\citet{ziyindeep} for details. An $m$-class classification task is defined as finding a function $f_\mathbf{w}: \mathbb{R}^d \to \mathbb{R}^m$, where $d$ is the input dimension, $m$ is the number of classes, and $\mathbf{w}$ denotes the parameters of $f$. We assume that the output $f(x)$ is normalized, and can be seen as the predicted probability of input $x$ being labeled in class $j$, i.e. $\Pr(j|x) = f_\mathbf{w} (x)_j$ and our goal is to maximize the log probability of the true label $j$:
\begin{equation}
    \max_\mathbf{w}\E[\log f_\mathbf{w}(x)_j]
\end{equation}

The gambler's loss involves adding an output neuron at the ``$0$-th dimension" to function as an abstention, or rejection, score. The new model, augmented with the rejection score, is trained through the \textit{gambler's loss function}:
\begin{equation}
    \max_\mathbf{w}  \sum_i^B \log \left[ f_{\mathbf{w}}(x_i)_{j(x_i)} + \frac{f_{\mathbf{w}}(x_i)_{0}}{\lambda}\right],
\end{equation}
where $1<\lambda\leq m$ is the hyperparameter of the loss function, interpolating between the cross-entropy loss and the gambler's loss and controlling the incentive for the model to abstain from prediction, with higher $\lambda$ encouraging abstention. It is shown that the augmented model learns to output abstention score $f_{\mathbf{w}}(x_i)_{0}$ that correlates well with the uncertainty about the input data point $x_i$, either when the data point $x_i$ has inherent uncertainty (i.e. its true label is not a delta function), or when $x_i$ is an out-of-distribution sample. This work studies the dynamic training aspects of the gambler's loss and utilizing these properties to pioneer a series of techniques that are robust in the presence of label noise. In particular, we argue that the gambler's loss is a noise-robust loss function.

\vspace{-2mm}
\section{Gambler's Loss is Robust to Label Noise}
\label{sec: gamblers loss}
\vspace{-1mm}

\begin{figure}[t]
    \centering
    
    \begin{subfigure}{0.46\linewidth}
    \includegraphics[trim=0 0 0 0, clip, width=\linewidth]{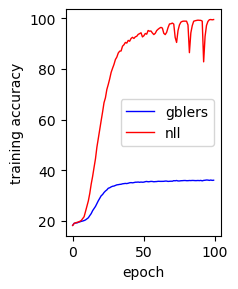}
    \vspace{-7mm}
    \caption{training acc.}
    \end{subfigure}
    \begin{subfigure}{0.42\linewidth}
    
    \includegraphics[trim=29.8 0 0 1, clip, width=\linewidth]{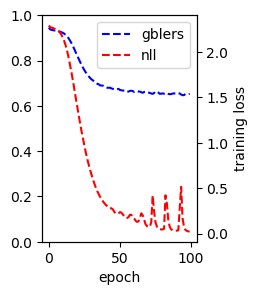}
    \vspace{-7mm}
    \caption{loss}
    \end{subfigure}

    \vspace{-3.5mm}
    \caption{\small{Training accuracy and training loss on an extremely corrupted MNIST dataset ($r=0.8$). In this case, $nll$ reaches $100\%$, meaning that the model has memorized all the corrupted data points.
    In contrast, training with gambler's loss reaches only $30\%$ training accuracy, resulting in a $60\%$ improvement in testing performance (at convergence, \textit{nll} loss reaches $19\%$ test accuracy, while gambler's loss obtains $76\%$). The gambler's loss is a more noise-robust loss function.}} 
    \label{fig:learnability}

\end{figure}

In this section, we examine binary classification problems in the presence of noisy labels. We begin by defining a bias-variance trade-off for learning with noisy labels (section~\ref{loss_s1}) and showing that the gambler's loss reduces generalization error (section~\ref{loss_s2}). Corroborated by the theory, we demonstrate three main effects of training with gambler's loss: 1) the gambler's loss automatically prunes part of the dataset (Figure~\ref{fig:learnability}); 2) the gambler's loss can differentiate between training data that is mislabeled and data that is cleanly labeled (Figure~\ref{fig:rejection score}); and 3) lower $\lambda$ can improve generalization performance (Figure~\ref{fig:robustness}). These theoretical and empirical findings suggest that training with the gambler's loss improves learning from noisy labels.

\vspace{-1mm}
\subsection{Bias-Variance Trade-off in Noisy Classification}
\label{loss_s1}
\vspace{-1mm}

The bias-variance trade-off is universal; it has been discovered in the regression setting that it plays a central role in understanding learning in the presence of label noise~\cite{krogh1992generalization, krogh1992simple, hastie2019surprises}. We first show that the loss function we are studying can be decomposed into generalized bias and variance terms; this suggests that, as in a regression problem, we might introduce regularization terms to improve generalization.

This section sets the notation and presents background for our theoretical analysis. Consider a learning task with $N$ input-targets pairs $\{(x_i, y_i)\}_{i=1,...,N}$ forming the test set. We assume that $(x_i y_i)$ are drawn i.i.d. from a joint distribution $p(x,y)= p(y|x)p(x)$. We also assume that for any given $x_i$, $y_i\in \{0, 1\}$ can be uniquely determined, so that $p(y|x) \in \{0, 1\}$. We also assume that the distribution of two classes are balanced, i.e. $p(y=1) = p(y=0) = 1/2$. We denote model outputs as $f(x_i):=f_i$. The \textit{empirical} generalization error $\ell_N[f]$ is defined as
\begin{align*}
    \ell_N[f] &=  -\sum_{y_i = 1} p(x_i, y_i) \log(f_i)-  \sum_{y_i = 0} p(x_i, y_i) \log (1 - f_i)\\
    &= -\frac{1}{N} \sum_{y_i = 1} y_i \log (f_i) - \frac{1}{N} \sum_{y_i = 0} (1 - y_i) \log (1 - f_i)
\end{align*}
which is the cross-entropy loss on the empirical test set. We assume that our model converges to the global minimum of the training objective since it has been proved that neural networks can find the global minimum easily~\cite{du2018gradient}. 

However, a problem with the binary label is that $\log 0$ diverges and the loss function diverges when a point is mislabeled, rendering the cross-entropy loss (also called $nll$ loss) very hard to analyze. To deal with this problem, we replace the binary label by slightly smoothed version $p(y_i|x=i) = (p, 1-p)$, where $p$ is the smoothing parameter and $1-p$ is perturbatively small~\citep{44903}. We will later take the limit $p\to y_i$ to make our analysis independent of the artificial smoothing we introduced. The optimal solution in this case is simply $f_i = \mathbb{E} [y_i] = p$, where the generalization error converges to
\begin{equation}
    \ell^* = -p\log p - (1-p)\log(1-p) = H(p);
\end{equation}
as $p\to 1$, the generalization error converges to $0$.

Now, we assume that label noise is present in the dataset such that each label is flipped to the other label with probability $r:= 1-a$, we assume that $r<0.5$. We define $r$ to be the \textit{corruption rate} and $a$ to be the \textit{clean rate}. The generalization error of our training dataset becomes
{\fontsize{9}{12}\selectfont
\begin{align}
\label{eq: training loss}
    \Tilde{\ell}[f] &= -\frac{1}{N} \sum_{\Tilde{y}_i = 1} \Tilde{y}_i \log (f_i) - \frac{1}{N} \sum_{\Tilde{y}_i = 0} (1 - \Tilde{y}_i) \log (1 - f_i)\\
    &= -\frac{1}{N}\left[ \sum_{y_i=1, \Tilde{y}_i = 1} \log (f_i)  + \sum_{y_i=1, \Tilde{y}_i = 0}  \log (1 - f_i)\right] \\
    &\quad -\frac{1}{N}\left[ \sum_{y_i=0, \Tilde{y}_i = 1} \log (f_i) + \sum_{y_i=0, \Tilde{y}_i = 0} \log (1 - f_i)\right]\\
    &= \ell_1[f] + \ell_0[f]\label{eq: training loss 2}
\end{align}
}where $\Tilde{y}$ denotes the new set of perturbed labels. Therefore, we have partitioned the original loss function into two separate loss functions, where $\ell_0$ is the loss for the data points whose original label is $0$, and likewise for $\ell_1$. The global optimum, as $N\to \infty$, for $\hat{\ell}_1$ is $H(p)$, the entropy of $p$. The generalization error of this solution is 
\begin{equation}\label{eq: raw generalization error}
    \ell(p,a) = aH(p) - (1-a)\left[p\log(1-p) + (1-p)\log p \right]
\end{equation}
where we have taken expectations over the noise. We observe a bias-variance trade-off, where the first term $H(p)$ denotes the variance in the original labels, while the second term is the bias introduced due to noise. As $a\to 1$, the noise disappears and we achieve perfect generalization where the training loss is the same as generalization loss.


\vspace{-1mm}
\subsection{Training and Robustness with the Gambler's Loss}
\label{loss_s2}
\vspace{-1mm}

The gambler's loss function was proposed by~\citet{ziyindeep} as a training method to learn an abstention mechanism. We propose to train, instead of on equation (\ref{eq: training loss}), but rather on the gambler's loss with hyperparameter $\lambda$:
\begin{align}
\label{eq: gamblers loss}
    \Tilde{\ell}[f] &= -\frac{1}{N} \sum_{\Tilde{y}_i = 1} \Tilde{y}_i \log \left(f_{i, 1} + \frac{f_{i, 0}}{\lambda} \right)\\
    &\quad- \frac{1}{N} \sum_{\Tilde{y}_i = 0} (1 - \Tilde{y}_i) \log \left(f_{i,2} + \frac{f_{i,0}}{\lambda}\right)
\end{align}
where we have rewritten $f_i \to f_{i, 1}$ and $1 - f_i \to f_{i, 2}$, and we have augmented the model with one more output dimension $f_0(x_i):= f_{i,0}$ denoting the rejection score. Notice that the gambler's loss requires the normalization condition:
\begin{equation}
    \underbrace{f_{i, 1}}_{\text{prediction on class 1}} + \underbrace{f_{i, 2}}_{\text{prediction on class 2}} + \underbrace{f_{i, 0}}_{\text{confidence score}} = 1
\end{equation}

\begin{figure}[t]
    \centering
    \vspace{-1mm}
    \includegraphics[trim=0 0 10 30, clip,width=1\linewidth]{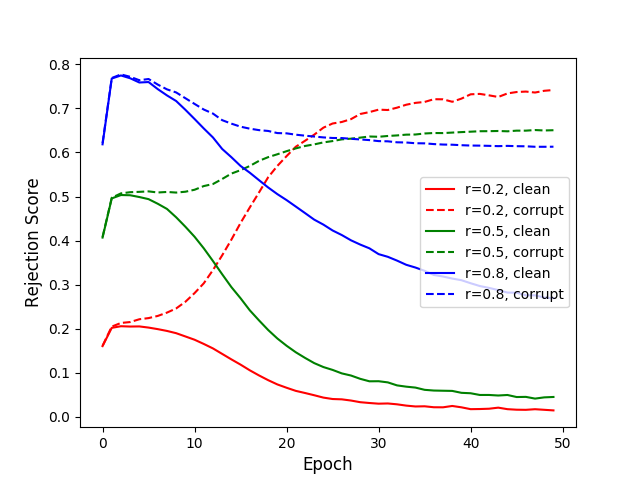}
    \vspace{-10mm}
    \caption{Rejection score on the clean and corrupted portions of the MNIST training set for $r=0.2, 0.5, 0.8$. We observe that our model learns rejection scores smaller than $0.5$ for data points with clean labels and learns larger rejection score for data points with corrupted labels, which agrees with our theoretical analysis. This implies that the model has learned \textit{not to learn} from the corrupted data points.}
    \label{fig:rejection score}
    
        \vspace{1.5mm}
    \includegraphics[trim=0 0 10 10, clip,width=0.8\linewidth]{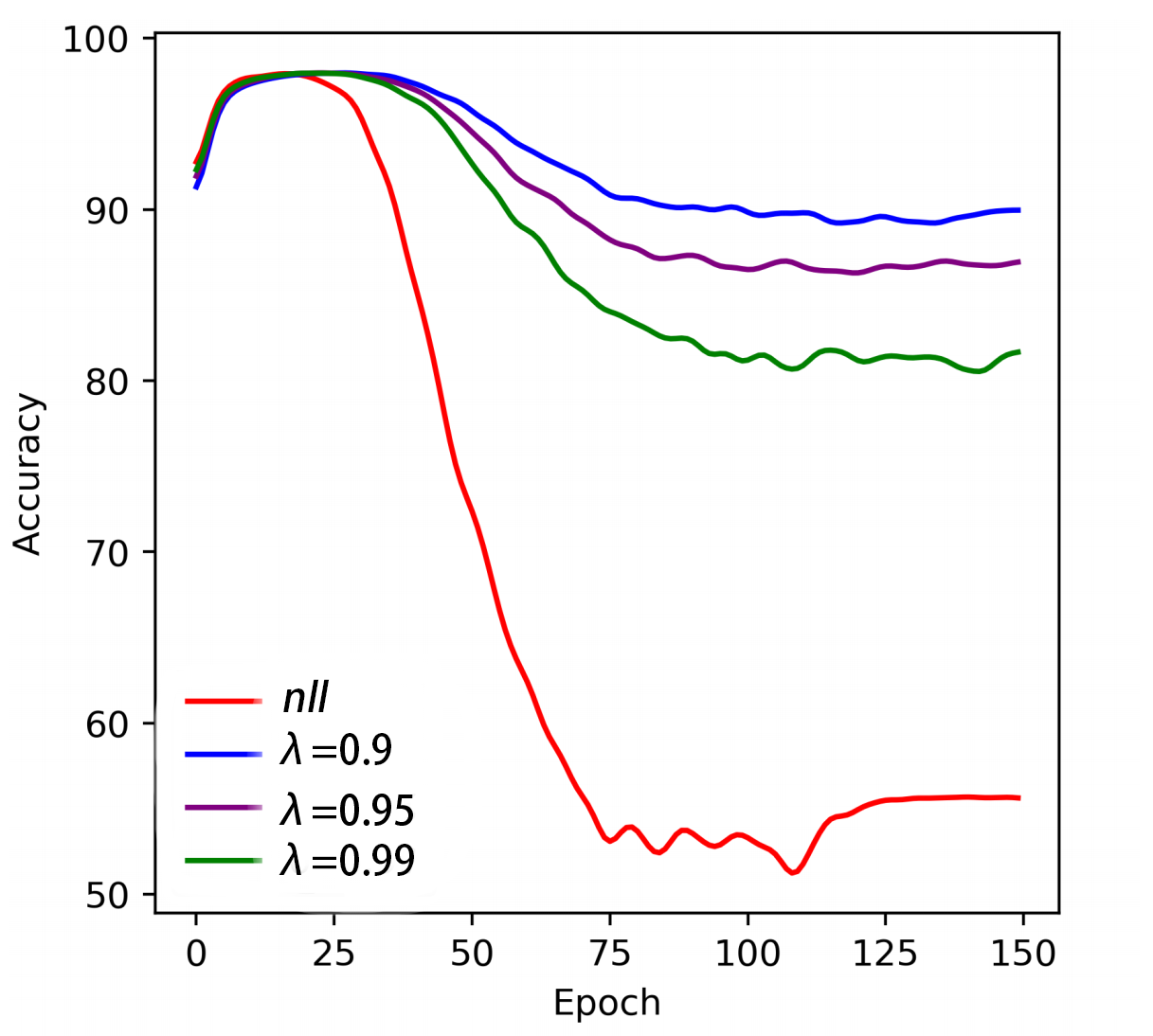}
    \vspace{-5mm}
    \caption{Robustness of gambler's loss to label corruption rate $r=0.5$ on MNIST, where ``nll" refers to the \textit{negative log-likelihood} loss. Changing $\lambda$ changes the training trajectory dramatically and lower $\lambda$ provides better robustness (higher accuracy).\vspace{-0mm}}
    \label{fig:robustness}
\end{figure}


To proceed, we define \textit{learnability} on a data point from the perspective of gambler's loss. We note that this definition is different but can be related to standard PAC-learnability~\cite{shalev2014understanding}.
\begin{definition}
    A data point $(x_i, y_i)$ is said to be \underline{not learnable} if the optimal solution on the gambler's loss of such point outputs $f_{i,0} = 1$. Otherwise, the point is \underline{learnable}.
\end{definition}
Since one category always predicts $1$, this prediction saturates the softmax layer of a neural network, making further learning on such a data point impossible as the gradient vanishes. If a model predicts $1$ as a rejection score, then it will abstain from assigning weight to any of the classes, thus avoiding learning from data point completely.

We now show that, when $\lambda < 1/p < 2$, then the points with $\max_j p(y=j) < p$ are not learnable.
\begin{theorem}
\label{theo: not learning}
    For a point $x_i$, with label $y_i=(p,\ 1-p)$ (assuming $p>1-p$)
    where $p$ denotes the probability that $y_i=1$, and if $ \lambda < 1/p$ then the optimal solution to the loss function
    $$\ell_i = - p\log\left(f_i + \frac{f_0}{\lambda}\right) - (1 -p)\log\left(1 - f_i - f_0 + \frac{f_0}{\lambda}\right)$$
    is given by 
    $$f_0 = 1,$$
    with $\ell_i =\log \lambda$, i.e. the model will predict $0$ on both classes.
\end{theorem}
See Appendix ~\ref{sec: gamblers loss proof} for the proof. Note that $\log(1/p)$ is roughly of similar magnitude to the entropy of $y_i$ given $x_i$. Therefore, if we want to prune part of the dataset that appears ``random'', we can choose $\log\lambda$ to be smaller than the entropy of that part of the dataset. To be more insightful, we have a control over the model complexity:
\begin{corollary}\label{theo: complexity}
    Let $X_\lambda$ be the subset of the dataset that are learnable at hyperparameter $\lambda$, and let $f^*(\cdot)$ be the optimal model trained on $X_\lambda$ using cross-entropy loss, then $H[f(x_i)|X_\lambda] \leq  H(\frac{1}{\lambda})$.
\end{corollary}


This result implies that the model will \textit{not learn} part of the dataset if it appears too random. See Figure~\ref{fig:learnability} for a demonstration of this effect, we see that training with the classical \textit{nll} loss memorizes all the data points at convergence, while gambler's loss selectively learns only $34\%$, resulting in a $56\%$ absolute performance improvement at testing.

The following theorem gives an expression for what an optimal model would predict for the learnable points.
\begin{theorem}
\label{theo: probability solution}
    Let $1<\lambda <m $, then the optimal solution to a learnable point $x_i$, whose label is $y_i$ and $\mathbb{E}[y_i] =p$, is
    \begin{equation}
        f_i^* = \frac{p\lambda -1 }{\lambda -1}.
    \end{equation}
\end{theorem}
This says that the optimal model will make the following prediction learnable point $x_i$, where $j = \arg\max f_j(x_i)$:
\begin{equation}
    \begin{cases}
    f_{0}(x_i) = 1 - \frac{p\lambda -1 }{\lambda -1} = \frac{\lambda(1-p)}{\lambda -1}, \\
    f_{j}(x_i) = \frac{p\lambda -1 }{\lambda -1}, \\
    f_{k}(x_i) = 0, &\text{for $k\neq j$}.
    \end{cases}
\end{equation}
By combining Theorem~\ref{theo: not learning} and Theorem~\ref{theo: probability solution}, we observe that the model will predict a higher rejection score $f_0(x)$ on the mislabeled points (close to $1$, since their conditional entropy is large), and lower rejection score on the correctly labeled data points (since their conditional entropy is small) in the training set. This is exactly what we observe in Figure~\ref{fig:rejection score}, where we plot the rejection scores on both clean and corrupted portions of the dataset, with noise corruption rates ranging from $r=0.2$ to $0.8$. We observe that our model learns rejection scores smaller than $0.5$ for data points with clean labels and learns larger rejection score for data points with corrupted labels. In other words, we are able to approximately \textit{filter} out the corrupted labels which can then be sent for relabeling in real-world scenarios.

\begin{figure*}[tbp]
    \begin{subfigure}{0.45\linewidth}
        \centering
        \includegraphics[width=1.05\linewidth]{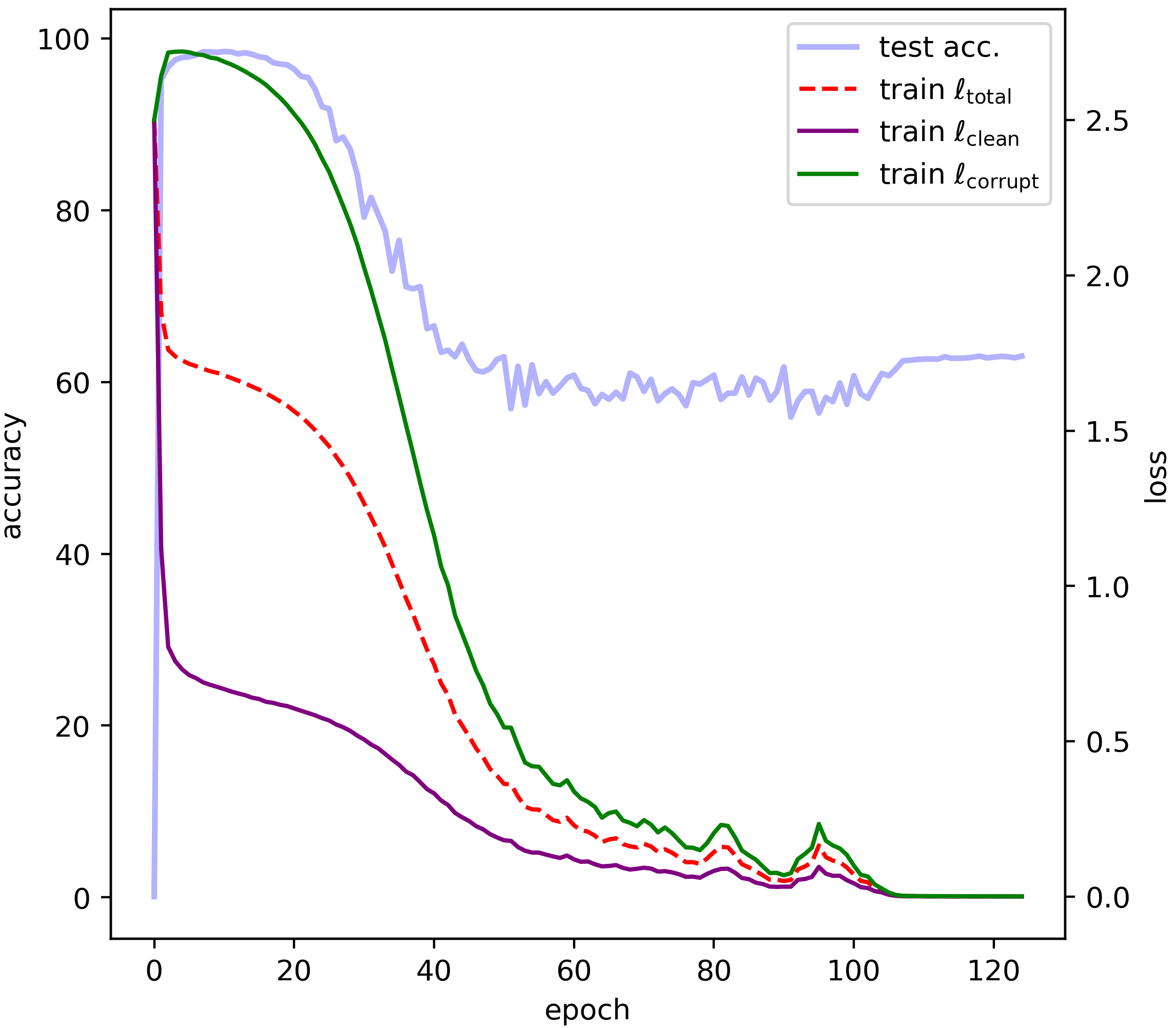}
        \vspace{-6mm}
        \caption{}
        
        \label{fig:demo_full}
    \end{subfigure}
    \hfill
    \begin{subfigure}{0.45\linewidth}
        \centering
        \includegraphics[width=1.05\textwidth]{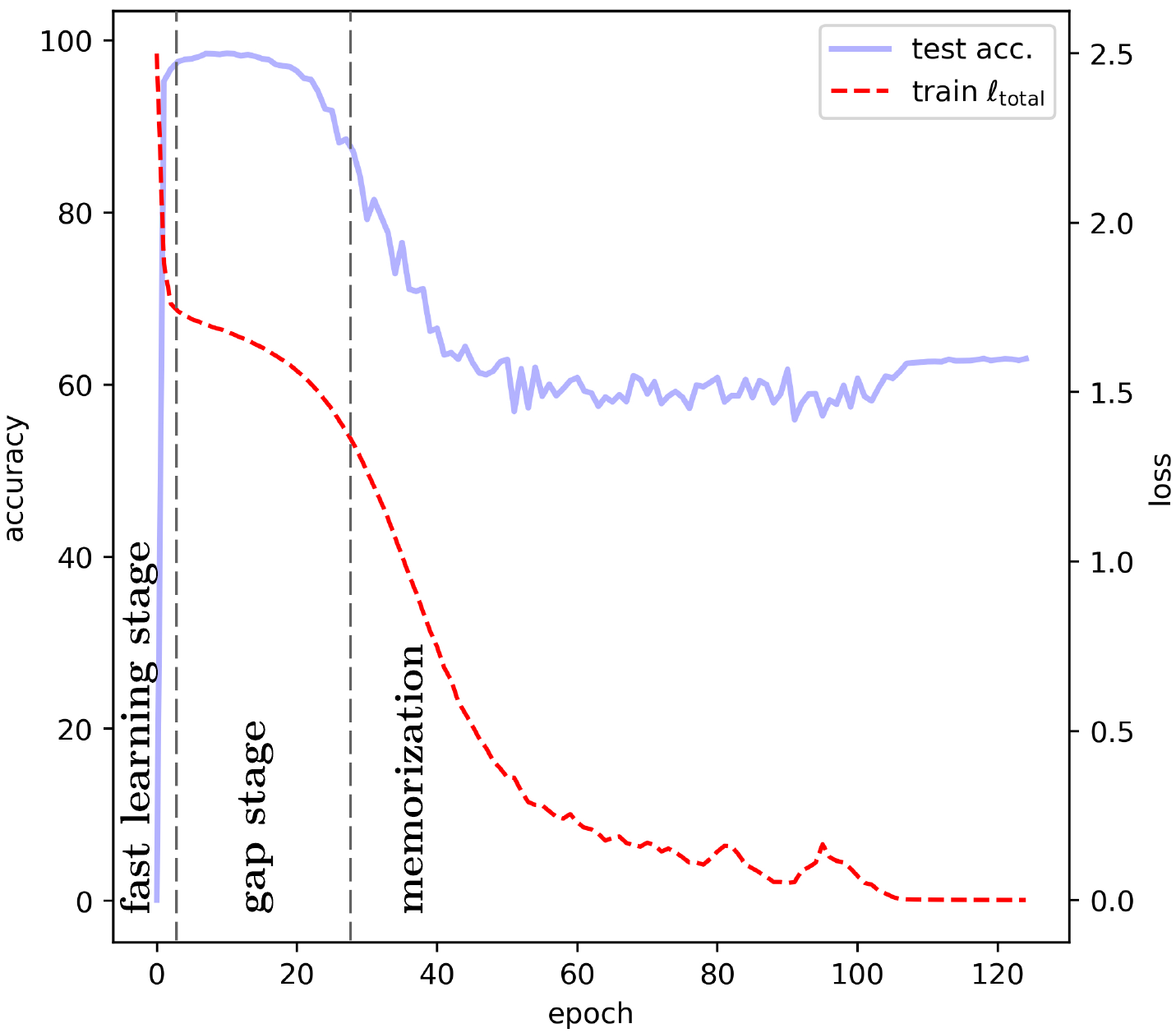}
        \vspace{-6mm}
        \caption{}
        \label{fig:demo_3-stage}
    \end{subfigure}
    \vspace{-3mm}
    \caption{\small{Different stages of training in the presence of label noise. (a) We plot $\ell_\mathrm{total}$ (total training loss), $\ell_\mathrm{clean}$ (training loss on the clean data), and $\ell_\mathrm{corrupt}$ (training loss on corrupt data), we see that at the plateau, there is a clear ``gap'' between the $\ell_\mathrm{clean}$ and $\ell_\mathrm{corrupt}$. (b) We highlight the $3$ training regimes: fast learning stage, gap stage, and the memorization stage. Experiment conducted on MNIST with corruption rate $0.5$.}}
    \label{fig:observation}
\end{figure*}

Since generalization error is reduced by training with the gambler's loss, we conclude that the gambler's loss is robust to the presence of noisy labels on learnable points.
\begin{theorem}
\label{theo: robustness}
    Let $\ell(p, a)$ be the generalization error achieved by the global minimum by the Kullbeck-Leibler's divergence with smoothing parameter $p$ and corruption rate $r=1-a$, given by equation (\ref{eq: raw generalization error}), and let $\ell_{\lambda}(p, a)$ be the generalization error achieved by training on the gambler's loss, where we make the prediction $(f_i^*, 1 - f_i^*)$ on each data point. Then, 
    \begin{equation}
        \lim_{p\to 1} \ell(p, a) - \ell_{\lambda}(p, a) = \log \left(\frac{\lambda}{\lambda - 1}\right)^{1-a} \geq 0,
    \end{equation}
    for $a\in (\frac{1}{2}, 1]$ and $\lambda > 1$. The equality is achieved when $a = 1$, i.e., when no noise is present.
\end{theorem}
Proof is given in Appendix~\ref{sec: robustness proof}. This shows that whenever noise is present, using gambler's loss will achieve better generalization than training using $nll$ loss, even if we are not aware of the corruption rate. Figure~\ref{fig:robustness} shows that lower $\lambda$ indeed results in better generalization in the presence of label noise. In addition, results in Table~\ref{tab: gambler-comparison} show that training with gambler's loss can improve over $nll$ loss when corruption is present.

\vspace{-3mm}
\section{Practical Extensions}
\label{sec: practical extension}
\vspace{-1.5mm}

While using the gambler's loss in isolation already gives strong theoretical guarantees and empirical performance in noisy label settings, we propose two practical extensions that further improve performance. The first is an analytical early stopping criterion that is designed to stop training before the memorization of noisy labels begins and hurts generalization. The second is a training heuristic that relieves the need for hyperparameter tuning and works well without requiring knowledge of the noise corruption rate.

\vspace{-1mm}
\subsection{An Early Stopping Criterion}
\label{sec: early stopping}
\vspace{-1mm}

\begin{figure}[t]
\centering
    \begin{subfigure}[b]{1\linewidth}
        \centering
        \includegraphics[width=0.85\textwidth]{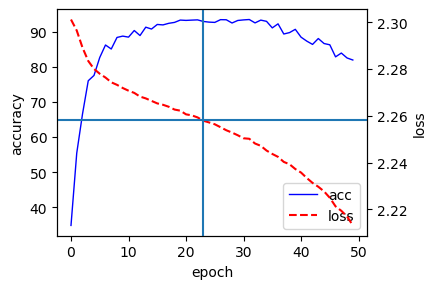}
        \vspace{-2mm}
        \caption{MNIST, CNN, $r = 0.80$}
        \label{fig: mnist0.85}
    \end{subfigure}
    \begin{subfigure}[b]{1\linewidth}
        \centering
        \includegraphics[width=0.85\textwidth]{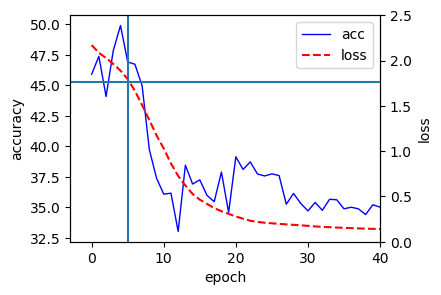}
        \vspace{-2mm}
        \caption{CIFAR10, ResNet18, $r = 0.85$}
        \label{fig: cifar0.85}
    \end{subfigure}
    \vspace{-8mm}
    \caption{\small{Early stopping results on MNIST and CIFAR10. The horizontal line is the predicted early stopping point based on theorem~\ref{theorem}. We see that our predicted early stopping point corresponds to the observed plateau where the testing accuracy (blue solid line) is at maximum, and right before test accuracy starts to decrease.\vspace{-0mm}}}
    \label{fig:earling stopping}
    \vspace{-3mm}
\end{figure}

We further propose an analytical early stopping criterion that allows us to perform early stopping with label noise \textit{without a validation set}. We begin by rewriting the generalization error we derived in equation (\ref{eq: training loss 2}) as a function of $\ell_1$:
{\fontsize{9}{12}\selectfont
\begin{align}
    N\ell_1 &= -\sum_{y_i=1, \Tilde{y}_i = 1} \log (f_i)  - \sum_{y_i=1, \Tilde{y}_i = 0}  \log (1 - f_i) \\
    &= - \sum_{y_i=1} \mathbb{I}_{\Tilde{y}_i = 1} \log (f_i)  - \sum_{y_i=1}  \mathbb{I}_{\Tilde{y}_i = 0}\log (1 - f_i) \\
    &\approx - \sum_{y_i=1} p(\Tilde{y}_i=1| y_i=1) \log (f_i)\\  
    &\ \ \ - \sum_{y_i=1}  p(\Tilde{y}_i=0| y_i=1)\log (1 - f_i) := N\bar{\ell}_1
\end{align}
}where, in the last line, we make a ``mean-field" assumption that the effect of the noise can be averaged over and replaced by the expected value of the indicators. This mean-field assumption is reasonable because it has been shown that learning in neural networks often proceeds at different stages, first learning low complexity functions and then proceeding to learn functions of higher complexity, with a random function being the most complex~\cite{nakkiran2019sgd}. Thus, while learning the simpler function of true features, we can understand the effect in a mean-field manner. We argue the validity of this assumption lies in its capability of predicting the point of early stopping accurately and thereby improving performance. We use the upper-bar mark to denote this mean-field solution (i.e. $\bar{\ell}_1$) and apply the same approximation to $\ell_0$, yielding $\bar{\ell}_0$.

Let $a = p(\Tilde{y}_i= y_i)$ denote the clean rate, we can derive an expression for the effective loss function $\bar{\ell}_1$ as follows:
\begin{theorem}
\label{theorem}
    The mean-field gambler's loss takes the form:
    \begin{equation}
        \bar{\ell} (\hat{p}) = - a \log \left( \hat{p} + \frac{1 - \hat{p}}{\lambda} \right) - (1 - a) \log \left( \frac{1 - \hat{p}}{\lambda} \right)
    \end{equation}
    which exhibits an optimal solution at training loss
    {\fontsize{9}{12}\selectfont
    \begin{align}
        \bar{\ell}^*(a, \lambda) &= \min_{p} \Tilde{\ell} (p) = - a \log a - (1 - a) \log \left( \frac{1 - a}{\lambda - 1} \right)\\
        &= H(a) + (1-a)\log(\lambda-1)
    \end{align}
    }which depends only on $a$ and $o$.
\end{theorem}
The proof is given in Appendix~\ref{sec: theorem proof}. Notice that this result extends directly without modification to the case of multiclass classification. Since the proof involves dealing with $\ell_0,\ \ell_1$ separately, and, as more classes are added, we only add terms such as $\ell_2,\ \ell_3,\ ...$ and so on. When $a<1$, the mean-field solution is greater than $0$, the global minimum, and this has the important implication that, when learning in the presence of label noise, a semi-stable solution at $\bar{\ell}^*>0$ exists, and is exhibited in the learning trajectory as a plateau around $\bar{\ell}$. After this plateau, the loss gradually converges to the global minimum at $0$ training loss. We refer to these three regimes with learning in the presence of noise as:

\vspace{-1mm}
1) \textbf{Fast Learning Stage}: The model \textit{quickly} learns the underlying mapping from data to clean labels; one observes rapid decrease in training loss and increases in test accuracy. 

\vspace{-1mm}
2) \textbf{Gap Stage}: This is the stage where the mean-field solution holds approximately. From Figure~\ref{fig:demo_full}, learning on the clean labels is almost complete ($\ell_{clean} \sim 0.5$) but training on noisy labels has not started yet ($\ell_\mathrm{corrupt} \sim 2.5$), and a large gap in training loss exists between the clean and corrupt part of the training set. Both the training loss and the test accuracy reach a plateau, and this is the time at which the generalization performance is the best.

\vspace{-1mm}
3) \textbf{Memorization}: This refers to the last regime when the model \textit{memorizes} these noisy labels and the train loss decreases slowly to $0$.

In addition to providing insights on the training trajectories in the presence of noisy labels, Theorem~\ref{theorem} also tells us that a network trained with the gambler's loss function with hyperparameter $\lambda$ on a symmetrically corrupted dataset with corruption rate $1 - a$ should have training loss around $\bar{\ell}^*(a, \lambda)$ during the gap stage, where \textit{generalization is the best}. This motivates using $\bar{\ell}^*(a, \lambda)$ as an \textit{early stopping criterion}. From the training plots in Figure~\ref{fig:earling stopping}, we see that the plateaus we hypothesize do indeed exist and our early stopping criterion accurately estimates the height of the plateau, thereby predicting close to the optimal early stopping point. In comparison with the standard early stopping technique by monitoring accuracy/loss on a validation set, we show that our proposed early stopping method is more accurate (Section~\ref{exp: early stopping}).
\begin{table}[t]
    \centering
    \begin{tabular}{c|cc}
         & large $\lambda$ &  small $\lambda$\\
         \hline
        learning speed  & $\uparrow$ & $\downarrow$ \\
       robustness  & $\downarrow$ & $\uparrow$
    \end{tabular}
    \vspace{-3mm}
    \caption{The learning speed and robustness trade-off that results from tuning the hyperparmeter $\lambda$ in the gambler's loss. In section~\ref{sec: schedule} we present an approach for automatically scheduling $\lambda$ to balance learning speed and robustness to noisy labels.}
    \label{tab:trade-off}
    \vspace{-5mm}
\end{table}

\vspace{-3mm}
\subsection{A Heuristic for Scheduling $\lambda$ Automatically}
\label{sec: schedule}
\vspace{-1.5mm}

\begin{table*}[!tbp]
\caption{\small{Robustness of the gambler's loss to noisy labels. We see that simply using gambler's loss on label noise settings improves on $nll$ loss. Furthermore, the scheduled gambler's loss achieves SOTA results as compared to other methods when corruption rate is unknown.}}
\label{tab: gambler-comparison}
\begin{center}
\vspace{-4.5mm}
\begin{tabular}{l|c|cc|ccc}
\hline
\hline
\multirow{2}{*}{\bf Dataset} & \multirow{2}{*}{\textit{nll} loss} & \multicolumn{2}{c|}{\textit{requires corruption rate}} & \multicolumn{3}{c}{\textit{agnostic to corruption rate}} \\ 
& & Coteaching+ & Gamblers (AES) & $L_q$ Loss & Gamblers & Gamblers + schedule  \\ 
\hline
\hline
MNIST $r=0.2$  & $95.1 \pm 0.2$ & $88.0\pm 0.1$ & $\mathbf{99.0\pm 0.1}$ & $\mathbf{98.9\pm 0.1}$ & $94.8\pm 0.3$ & $98.7\pm 0.1$ \\

MNIST $r=0.5$  & $74.1\pm 1.0$  & $87.2\pm 0.2$ & $\mathbf{98.4\pm 0.2}$ & $97.8\pm 0.2$ & $80.8\pm 0.5$ & $\mathbf{98.2\pm 0.1}$\\

MNIST $r=0.65$ & $54.5\pm 0.5$  & $86.1\pm 0.2$ & $\mathbf{97.6\pm 0.3}$& $82.5\pm 0.6$ & $66.3\pm 0.6$ & $\mathbf{97.9\pm 0.1}$ \\

MNIST $r=0.8$  & $21.7 \pm 0.6$ & $63.8\pm 0.1$ & $\mathbf{95.0\pm 0.5}$& $39.3\pm 1.3$ & $46.7\pm 0.8$ & $\mathbf{95.0\pm 0.1}$  \\


\hline


IMDB $r=0.15$   & $72.7\pm 0.5$ & $\mathbf{77.6\pm 0.4}$ & $75.4\pm 1.0$ & $74.1\pm 1.0$ & $72.7\pm 0.2$ & $\mathbf{75.0\pm 0.6}$\\

IMDB $r=0.2$    & $69.4\pm 0.7$ & $\mathbf{75.0\pm 0.3}$ & $71.6\pm 2.9$ & $70.8\pm 0.8$ & $69.0\pm 0.7$ & $\mathbf{72.8\pm 0.5}$ \\

IMDB $r=0.25$   & $65.4\pm 0.8$ & $\mathbf{72.3\pm 0.4}$ & $70.5\pm 5.8$ & $68.7\pm 0.7$ & $64.8\pm 0.7$ & $\mathbf{68.9\pm 0.5}$ \\

IMDB $r=0.3$    & $54.6\pm 0.1$ & $\mathbf{67.3\pm 0.3}$ & $54.5\pm 2.3$ & $61.7\pm 0.5$ & $59.8\pm 0.9$ & $\mathbf{62.2\pm 2.2}$ \\

\hline

CIFAR10 $r=0.2$  & $52.7 \pm 0.1$ & $\mathbf{52.2\pm 0.1}$ & $52.1\pm 0.3$ & $52.3\pm 0.4$ & $\mathbf{52.9\pm 0.3}$ & $\mathbf{52.8\pm 0.4}$ \\

CIFAR10 $r=0.5$  & $41.1\pm 0.0$  & $47.4\pm 0.0$ & $\mathbf{48.3\pm 0.4}$ & $46.6\pm 0.3$ & $40.6\pm 0.3$ & $\mathbf{51.9\pm 0.2}$ \\

CIFAR10 $r=0.65$ & $31.8\pm 0.4$  & $42.5\pm 0.1$ & $\mathbf{43.2\pm 0.8}$ & $34.5\pm 0.5$  & $30.8\pm 0.3$ & $\mathbf{45.6\pm 0.3}$ \\

CIFAR10 $r=0.8$  & $10.6 \pm 0.5$ & $9.9\pm 0.0$ & $\mathbf{29.7\pm 1.0}$ & $20.1\pm 0.1$ & $18.1\pm 0.2$ & $\mathbf{25.4\pm 0.7}$  \\

\end{tabular}
\vspace{-4mm}
\end{center}
\end{table*}
\begin{table}[!tbp]
\vspace{-2mm}
\caption{{Early stopping experiment. The number in parenthesis is the average number of epochs until early stopping. Gambler's loss with early stopping (AES) stops faster and at better accuracy.}}
\vspace{-1.5mm}
\label{tab:earlystopping}
\begin{center}

\begin{tabular}{l|cc}
\hline \hline
{\bf Dataset} & VES & Gamblers AES \\ 
\hline
\hline
MN $r=0.2$      & $95.1\pm 0.4\ (95)$  & $\mathbf{98.8\pm 0.1\ (17)}$ \\
MN $r=0.5$      & $79.7\pm 2.0\ (115)$ & $\mathbf{98.0\pm 0.0\ (18)}$ \\
MN $r=0.8$      & $21.1\pm 0.1\ (117)$ & $\mathbf{93.5\pm 0.3\ (15)}$\\
MN $r=0.85$     & $15.0\pm 1.1\ (110)$ & $\mathbf{85.2\pm 1.7\ (13)}$\\
\hline
IMDB $r=0.1$    & $71.0\pm 0.5\ (46)$ & $\mathbf{74.0\pm 0.4\ (10)}$\\
IMDB $r=0.2$    & $59.6\pm 0.8\ (8)$  & $\mathbf{65.3\pm 0.7\ (12)}$\\
IMDB $r=0.3$    & $51.1\pm 0.2\ (8)$  & $\mathbf{58.2\pm 0.3\ (12)}$\\

\end{tabular}
\vspace{-2em}
\end{center}
\end{table}

While the above section presents an effective guideline for early stopping in the presence label noise, it still requires tuning the hyperparameter $\lambda$. In practice, choosing for the optimal $\lambda$ is not straightforward and requires special tuning for different tasks. In this section, we present a heuristic that eliminates the need for tuning $\lambda$. It also carries the important benefit of not requiring knowledge of the label corruption rate.

This section is based on the gambling analogy \cite{CoverInformationTheory, Portfoliotheory, UniversalPortfolio} and the following two properties of training on gambler's loss: (1) larger $\lambda$ encourages feature learning and smaller $\lambda$ slows down learning \cite{ziyindeep}; (2) smaller $\lambda$ encourages robustness and larger $\lambda$ provides less robustness (this paper). This means that there is a trade-off between the speed of feature learning and robustness when using gambler's loss. As a result, it would be ideal to balance both aspects and set $\lambda$ to achieve a better trade-off between robustness and feature learning. See Table~\ref{tab:trade-off} for a summary of these trade-offs.

Recall that for true label $j$, the gambler's loss is given by
\begin{equation}
    {\ell}_j = -\log \left(f_{j}(x_i) + \frac{f_{0}(x_i)}{\lambda} \right).
\end{equation}
This loss can decrease in two ways: (1) (\textit{trivial learning}) one may trivially decrease the loss by increasing $f_0(x_i)$ which is the rejection score. Since $f_0$ is present in every category, this does not correspond to learning; (2) one may increase the output on the true label $f_j(x_i)$, which corresponds to actual learning. While lower $\lambda$ gives better robustness to label noise, it also encourages trivial learning. In fact, as shown in Theorem~\ref{theo: not learning}, choosing too small $\lambda$ leads to a trivial solution with a loss of $\log\lambda$. Therefore, one is motivated to choose the lowest $\lambda$ such that normal learning can still occur. More importantly, since different data points are learned at potentially different speeds~\citep{10.1145/1553374.1553380}, we propose a rule to automatically set $\lambda_i$ adaptively as a function of each data point (rather than a general $\lambda$). For a data point $(x_i)$, let $f(x_i)\in \mathbb{R}^{m+1}$ denote the predicted probability. We choose $\lambda_i$ to be
\begin{equation}
\label{lambda_v1}
\boxed{
    \lambda_i = \frac{(\sum_{j=1}^m f_j(x_i))^2}{\sum_{k=1}^m f_k(x_i)^2}}.
\end{equation}
Firstly, as a sanity check, the Cauchy-Schwarz inequality tells us that $1\leq \lambda_i \leq m$, which is in the well-defined range for gambler's loss. The choice for $\lambda_i$ in equation (\ref{lambda_v1}) comes from the fact that we can view the classification problem as a horse race betting problem. In the gambler's loss analogy, $\lambda_i$ represents the return on a correct bet~\cite{ziyindeep}. As the model trains, the optimal $\lambda_i$ will tend to decrease as the model gains confidence on the data in the training set and is less likely to resort to trivial learning. Thus, intuitively, we would like to decrease $\lambda_i$ as the model grows more confident. To achieve this, we examine the gain of the gambler from a single round of betting:
\begin{equation}
    S_i = \sum_{j=1}^{m}( \lambda_i p_j f_j ) + f_0,
\end{equation}
where we write $f_j := f_j(x_i)$ for concision. Greater model confidence corresponds to greater expected gain on the part of the gambler since the gambler will consolidate bets on certain classes as certainty increases. Therefore, to achieve a $\lambda_i$ appropriate for the current gain, we set $\lambda_i$ such that the expected gain of the gambler is constant. Since our metaphorical starts with $1$ units of currency to gamble, we naturally choose $S_i=1$ as our constant:
\begin{equation}
    \lambda_i = \mathbb{E}_{\text{gambler}}\bigg[\frac{1 - f_0}{\sum_{j=1}^{m} p_j f_j }\bigg].
\end{equation}
To the gambler, the only unknown is $p_j$, but we can recover the gambler's expectation through his bets: $\mathbb{E}_{\text{gambler}}[p_j]=f_j / \sum_{k=1}^m f_k$. Thus, we obtain:
\begin{equation}
\label{rule}
    \lambda_i = \frac{(1 - f_0)^2}{\sum_{k=1}^{m} f_k^2 } = \frac{(\sum_{j=1}^{m} f_j)^2}{\sum_{k=1}^{m} f_k^2 }
\end{equation}
In section~\ref{exp: automatic scheduling}, we extensively study the performance of automatic scheduling using $\lambda_i$ and we show that it achieves SOTA results on three datasets.

In Appendix~\ref{sec: alternative scheduling rule}, we derive and discuss an alternative scheduling rule $\lambda_{exp}$ by directly setting the doubling rate $\log S$ (i.e. the loss function we are optimizing over) to $0$ (i.e. its minimum value). This gives us
\begin{equation}
    \lambda_{exp}(x_i) = \exp \left[-\frac{\sum_{j=1}^{m}  f_j \log f_j }{\sum_{k=1}^{m}  f_k}\right].
\end{equation}
By Jensen's inequality, 
\begin{equation}
    \underbrace{\frac{(\sum_{j=1}^{m}  f_j)^2}{\sum_{k=1}^{m}  f_k^2}}_{\lambda_{i}} \leq \frac{\sum_{j=1}^{m}  f_j}{\sum_{k=1}^{m}  f_k^2} \leq \underbrace{\exp \left[- \frac{\sum_{j=1}^{m}  f_j \log f_j }{\sum_{k=1}^{m}  f_k}\right]}_{\lambda_{exp}},
\end{equation}
showing that $\lambda_{exp}$ encourages learning more at the expense of robustness. If better training speed is desired, we expect $\lambda_{exp}$ to perform better. While our experiments focus on showcasing the effectiveness of $\lambda_{i}$, we discuss the performance of both strategies in Appendix~\ref{sec: alternative scheduling rule}.


\vspace{-3mm}
\section{Benchmark Experiments}
\label{sec: experiments}
\vspace{-1.5mm}

In this section, we experiment with the proposed methods under standard label noise settings. We first show that our proposed early stopping criterion (section~\ref{sec: early stopping}) stops at a better point as compared to classical early stopping methods based on validation sets. Next, we show that dynamic scheduling using $\lambda_{i}$ (section~\ref{sec: schedule}) achieves state-of-the-art results as compared to existing baselines.

\vspace{-3mm}
\subsection{Early Stopping Criterion}
\label{exp: early stopping}
\vspace{-1.5mm}

We split $6000$ images from the training set to make a validation set, and we early stop when the validation accuracy stops to increase for $5$ consecutive epochs. There are also a few other validation-based early stopping criteria, but they are shown to perform similarly~\citep{prechelt1998early}. We refer to early stopping techniques based on monitoring the validation loss as \textbf{VES} (validation early stopping) and call our method \textbf{AES} (analytical early stopping). We fix $\lambda=9.99$ when training our models and collect the results in Table~\ref{tab:earlystopping}. On MNIST, we see that our proposed method significantly outperforms the baseline early stopping methods both by testing performance (up to $70\%$ in absolute accuracy) and training time ($10$ times faster); We also conduct experiments on the IMDB dataset~\citep{maas-EtAl:2011:ACL-HLT2011}, which is a standard NLP sentiment analysis binary classification task. We use a standard LSTM with a hidden dimension of $256$ and $300$-dimensional pretrained GloVe word embeddings~\citep{pennington2014glove}. Again, we notice that AES consistently improves on early stopping on a validation set (by about $2-7\%$ in absolute accuracy). We hypothesize that the small sizes of the validation set result in a large variance of the early stopping estimates. This problem becomes more serious when label noise is present. On the other hand, AES does not require estimation on a small validation set and is more accurate for early stopping.

\vspace{-2mm}
\subsection{Automatic Scheduling}
\label{exp: automatic scheduling}
\vspace{-1mm}

In this section, we demonstrate that the proposed scheduling method achieves very strong performance when compared to other benchmarks, which include the generalized cross-entropy loss ($L_q$) \cite{zhang2018generalized} and Coteaching+ \cite{yu2019does}. Generalized cross-entropy loss serves as a direct comparison to our scheduling method: it is the current SOTA method for noisy label classification that is agnostic to the corruption rate and modifies only the loss function, two qualities shared by our method. We set the hyperparameter $q=0.7$ for $L_q$ following the experiments in~\cite{zhang2018generalized}. Meanwhile, Coteaching+, the SOTA method when the corruption rate is known, introduces a novel data pruning method. In our comparison, we give Coteaching+ the true corruption rate $r$, but we note that its performance is likely to drop when $r$ is unknown and has to be estimated beforehand. We perform experiments on $3$ datasets, ranging from standard image classification tasks (MNIST: $10-$class; CIFAR-10: $10-$class) to text classification tasks using LSTM with attention and pretrained GloVe word embedding (IMDB: $2-$class). For the IMDB dataset, we use the LaProp optimizer \cite{ziyin2020laprop}. We note that training with LaProp is faster and stabler than using Adam \cite{journals/corr/KingmaB14_adam}.

From the results in Table~\ref{tab: gambler-comparison}, we see that automatic scheduling outperforms $L_q$ in $11$ out of $12$ categories in a statistically significant way. More importantly, we see \textit{larger margins of improvement as the noise rate increases}. $L_q$ is only better than scheduled gambler's loss on MNIST at the lowest corruption rate ($r=0.1$) and only by $0.2\%$ accuracy. Furthermore, the scheduled gambler's loss also outperforms Coteaching+ on 2 out of 3 datasets we compare on ($8$ categories out of $12$), while using only \textit{one half of the training time} and \textit{not requiring knowledge of the true corruption rate}. For CIFAR-10 and MNIST, the gambler's loss is especially strong when the noise rate is extreme. For example, when $r=0.8$, the scheduled gambler's loss significantly outperforms Coteaching+ by $30\%$ in absolute accuracy on MNIST, and by $15\%$ in absolute accuracy on CIFAR-10.

\vspace{-3mm}
\section{Conclusion}
\vspace{-1.5mm}

In this paper, we demonstrated how a theoretically motivated study of learning dynamics translates directly to the invention of new effective algorithms. In particular, we showed that the gambler's loss function features a unique training trajectory that makes it particularly suitable for robust learning from noisy labels and improving the generalization of existing classifications models. We also presented two practical extensions of the gambler's loss that further increase its effectiveness in combating label noise: (1) an early stopping criterion that can be used to accelerate training and improve generalization when the corruption rate is known, and (2) a heuristic for setting hyperparameters which does not require knowledge of the noise corruption rate. Our proposed methods achieve the state-of-the-art results on three datasets when compared to existing label noise methods.

\clearpage

{\small

\bibliographystyle{icml2020}
}


\onecolumn 
\appendix

\section{Proof of Theorem~\ref{theo: not learning} and Theorem~\ref{theo: probability solution}}
\label{sec: gamblers loss proof}
Taking the gambler's analogy, this theorem simply means that a gambler betting randomly will not make money, and so the better strategy is to reserve money in the pocket. 
Let $\lambda$ be the gambler's hyperparameter. Let $\hat{p}$ be the predicted probability on the true label $y$, and let $\hat{k}$ denote the prediction made on all the wrong classes added altogether, $\hat{l}$ be the predicted confidence score by the gambler's loss. By definition of a probability distribution, we have $\hat{p} + \hat{k} + \hat{l} = 1$.

We first show that $\hat{k}=0$. Intuitively speaking, this simply means that a gambler betting randomly will not make money, and so the better strategy is to reserve money in the pocket, and so it suffices to show that for any solution $\hat{p}, \hat{k}, \hat{l}$, the solution $\hat{p}' = \hat{p}, \hat{k}' = 0, \hat{l}' = \hat{l} + \hat{k}$ achieves better or equal doubling rate. For a mislabeled point (we drop $\hat{\cdot}$), the loss is $\log(\frac{k}{M} + \frac{l}{\lambda})$
but $M>\lambda$, and so $\log(\frac{k}{M} + \frac{l}{\lambda}) < \log(\frac{k + l}{\lambda})$, and we have that optimal solution always have $\hat{k} = 0$.

Now, we find the optimal solution to 
\begin{equation}
        \Tilde{\ell} (\hat{p}) = -p \log \left( \hat{p} - \frac{1 - \hat{p}}{\lambda} \right) - (1 - p) \log \left( \frac{1 - \hat{p}}{\lambda} \right)
\end{equation}
by taking the derivative with respect to p:
\begin{equation}
        \frac{\partial \Tilde{\ell}}{\partial \hat{p}} (\hat{p}) = - p \frac{\lambda-1}{(\lambda-1) \hat{p}+1} - (1-p) \frac{-1}{1-\hat{p}}
\end{equation}
and then setting it equal to 0
\begin{equation}
        \frac{\partial \Tilde{\ell}}{\partial \hat{p}} (\hat{p}) = -p \frac{\lambda-1}{(\lambda-1) \hat{p}+1} - (1-p) \frac{-1}{1-\hat{p}}=0
\end{equation}
is the $\hat{p}_{optimal}$:
\begin{equation}
        \hat{p}_{optimal}=\frac{p \lambda - 1}{\lambda - 1}
\end{equation}
and notice that $\hat{p} = 0$ if $p<\frac{1}{\lambda}$.

\section{Proof of Corollary~\ref{theo: complexity}}
Consider a data point $(x, y)$, WLOG, assume $1$ is the correct label, then $p:= p(y=1) \geq \frac{1}{\lambda}$, and this has at most $H(\frac{1}{\lambda})$, and so the output of the optimal model would have the same entropy, since the optimal prediction is proportional to $p$.

\section{Proof of Theorem~\ref{theo: robustness}}
\label{sec: robustness proof}
We want to show:
\begin{align}
    \lim_{p\to 1} \ell(p,a) - \ell_\lambda(p^*,a) = \lim_{p\to 1} aH(p) &- (1-a)\left[p\log(1-p) + (1-p)\log p \right] \\
    &- aH(p^*) + (1-a)\left[p^*\log(1-p^*) + (1-p^*)\log p^* \right] \geq 0
\end{align}
where $p^*=\frac{p \lambda - 1}{\lambda - 1}$ is given by Theorem~\ref{theo: probability solution}. Plug in to get
\begin{align}
    \lim_{p\to 1} \ell(p,a) - \ell_\lambda(p^*,a) &= \lim_{p\to 1} - (1-a)\left[p\log(1-p) + (1-p)\log p \right] + (1-a)\left[p\log(1-p^*) + (1-p)\log p^* \right]  \\
    & = \lim_{p\to 1} (1-a)\left[p\log\left(\frac{1 - p^*}{1-p}\right) + (1-p)\log \frac{p^*}{p} \right]\\
    &=  \lim_{p\to 1} (1-a)\left[p\log\left(\frac{1 - p^*}{1-p}\right)\right]
\end{align}
we can apply L'Hopital's rule to obtain
\begin{align}
    \lim_{p\to 1} \ell(p,a) - \ell_\lambda(p^*,a) = (1-a)\log\left(\frac{\lambda}{\lambda-1}\right) = \log\left(\frac{\lambda}{\lambda-1}\right)^{1-a} \geq 0
\end{align}
where the inequality follows from the fact that $1\leq \lambda \leq 2 $, and this is the desired result.

\subsection{A little further derivation...}
We might also obtain a perturbative result when $1 - p \ll a$ but is finite:
\begin{corollary}
    Let $1-p \ll a$, then
    \begin{equation}
        \ell(p, a) - \ell_{\lambda}(p, a) = \frac{(1-a)p}{\lambda-1} + O\left((1-p)^2\right) \geq 0
    \end{equation}
    for $a\in (\frac{1}{2}, 1]$ and $\lambda > 1$. The equality is achieved when $a = 1$, i.e., when no noise is present.
\end{corollary}

\section{Proof of Theorem~\ref{theorem}}
\label{sec: theorem proof}
To do this, one simply has to notice that theorem~\ref{theo: probability solution} applies with $p=a$, and we can plug in the optimal solution:
\begin{equation}
        \hat{p}_{optimal}=\frac{a \lambda - 1}{\lambda - 1}
\end{equation}
then plugging into the original equation [9]:
\begin{equation}
        \Tilde{\ell}^*(a, \lambda) = \min_{p} \Tilde{\ell} (p) = - a \log a - (1 - \epsilon) \log \left( \frac{1 - a}{\lambda - 1} \right)
\end{equation}

\section{An alternative Scheduling Rule}\label{sec: alternative scheduling rule}
Since adapting $\lambda_i$ might result in a small value for $\lambda_i$, rule (\ref{rule}) might slow down the training speed.

\begin{figure}
    \centering
    \includegraphics[trim=0 0 0 25 , clip, width=0.5\linewidth]{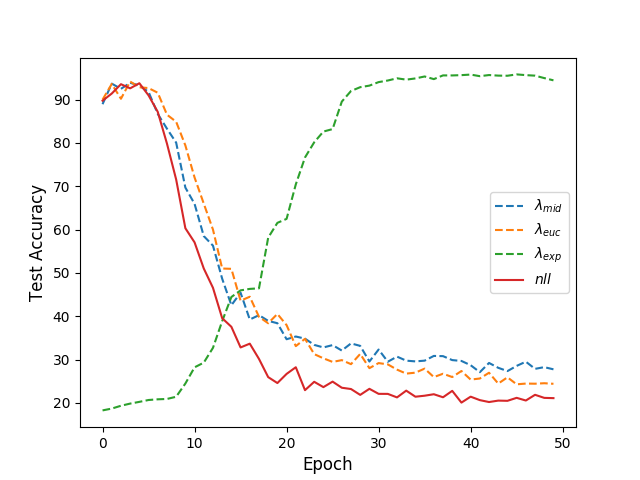}
    \vspace{-5mm}
    \caption{Training trajectories of different.}
    \label{fig:schemes}
\end{figure}

Since the doubling rate $\log S$ is the loss function we are optimizing over, another way to obtain $\lambda$ is to set  $\log S$ to be $0$ (i.e. its minimum value). This gives us
\begin{equation}
    \sum_{j=1}^{m}  p_j \log\left( \lambda f_j  + f_0 \right) = 0
\end{equation}
and again replacing $p_j$ with $f_j/\sum_{k=1}^m f_k$, we obtain
\begin{align}
    \mathbb{E}[\log S] &= \sum_{j=1}^{m}  f_j \log (\lambda f_j  + f_0 ) \\
    &= \log\lambda + \left[\sum_{j=1}^{m}  \frac{f_j}{\sum_{k=1}^m f_k} \log \left(f_j  + \frac{f_0}{\lambda} \right)\right]\\
    & \approx \log\lambda + \left[\sum_{j=1}^{m}  \frac{f_j}{\sum_{k=1}^m f_k} \log \left(f_j  \right)\right] = 0
\end{align}
where we assumed that $f_0 \ll 1$. Rearranging terms gives us
\begin{equation}
    \lambda_{exp}(x_i) = \exp \left[-\frac{\sum_{j=1}^{m}  f_j \log f_j }{\sum_{k=1}^{m}  f_k}\right]
\end{equation}
which is well in the range $[1, m]$. We give it a subscript $exp$ ($\lambda_{exp}$) to denote that it is different from the previous euclidean-style scheduling of $\lambda$ ($\lambda_{euc}$). This scheduling rule is quite aesthetically appealing since the term on the right takes the form of an entropy. We can apply Jensen's inequality to obtain the relationship between $\lambda_{exp}$ and $\lambda_{euc}$:
\begin{equation}
    \underbrace{\frac{(\sum_{j=1}^{m}  f_j)^2}{\sum_{k=1}^{m}  f_k^2}}_{\lambda_{euc}} \leq \underbrace{\frac{\sum_{j=1}^{m}  f_j}{\sum_{k=1}^{m}  f_k^2}}_{\lambda_{mid}} \leq \underbrace{\exp \left[- \frac{\sum_{j=1}^{m}  f_j \log f_j }{\sum_{k=1}^{m}  f_k}\right]}_{\lambda_{exp}},
\end{equation}
showing that $\lambda_{exp}$ encourages learning more than $\lambda_{euc}$ while $\lambda_{euc}$ provides stronger robustness. It is possible that $\lambda_{euc}$ might slows down training, and if better training speed is desired, we expect using $\lambda_{exp}$ is better. See Figure~\ref{fig:schemes}; we show the training trajectory of three different $\lambda$ schemes vs. that of $nll$ loss. We see that, as expected, all three schemes results in better performance than $nll$ at convergence. The $\lambda_{euc}$ scheme offers stronger robustness while reducing the training speed, while the other two schemes learns faster. While our experiments focuses on showing the effectiveness of $\lambda_{euc}$, we encourage the practitioners to also try out $\lambda_{exp}$ when necessary. We also expected $\lambda_{exp}$ to be helpful for standard image classification tasks when the noise rate is small but unknown. 

\section{Assymetric Noise Experiment}

The assymetric noise we experiment with is called the pairflip type of noise, which is defined in \cite{yu2019does}. See Table~\ref{tab: assymetric}. Throughout our experiments, we used a uniform $0.001$ learning rate with batch sizes of $128$. Our MNIST and CIFAR10 models were trained with Adam optimizers, while IMDB was trained with an experimental optimizer. For MNIST and CIFAR10, we used a 2-layer CNN, MNIST with 2 fully connected layers and CIFAR10 with 3. Accuracy was recorded after a previously set number of epochs (except under the early stopping criterion experiments). In IMDB, we used a single-layer LSTM. In MNIST, the model was trained for $50$ epochs, using auto-scheduled gambler's loss throughout. In CIFAR10 and IMDB, the model was trained for $100$ epochs, with the first $10$ epochs run with regular cross-entropy loss. We note that the proposed method also achieves the SOTA results on assymetric noise.

\begin{table*}[!tbp]
\caption{\small{Robustness of the gambler's loss to assymetric noise.}}
\label{tab: assymetric}
\begin{center}
\begin{tabular}{l|c|cccc}
\hline
\hline
{\bf Dataset} & {\textit{nll} loss} 
& Coteaching+  & $L_q$ Loss & Gamblers & Gamblers + schedule  \\ 
\hline
\hline
MNIST $r=0.2$  & $83.8\pm 0.9$ & $87.9\pm 0.2$  & $98.7\pm 0.1$ & $82.9\pm 0.7$ & $\mathbf{98.9\pm 0.1}$ \\

MNIST $r=0.3$  & $72.1\pm 1.0$ & $87.6\pm 0.3$ &  $93.4\pm 0.5$ & $73.0\pm 0.8$ & $\mathbf{98.6\pm 0.1}$\\

MNIST $r=0.45$ & $55.3\pm 0.5$ & $\mathbf{74.5\pm 1.0}$ &  $55.5\pm 1.4$ & $55.0\pm 1.2$ & $\mathbf{72.8\pm 3.0}$ \\


\hline

CIFAR10 $r=0.2$  & $52.7\pm 0.5$ & $52.2\pm 0.1$  & $50.2\pm 0.3$ & $52.4\pm 0.3$ & $\mathbf{56.4\pm 0.3}$ \\

CIFAR10 $r=0.3$  & $47.7\pm 0.8$ & $49.9\pm 0.1$ &  $46.7\pm 0.3$ & $47.5\pm 0.6$ & $\mathbf{54.8\pm 0.3}$\\

CIFAR10 $r=0.45$ & $34.5\pm 1.0$ & $38.9\pm 0.2$ &  $34.6\pm 0.8$ & $24.6\pm 0.8$ & $\mathbf{46.7\pm 1.0}$ \\

\hline







\end{tabular}
\vspace{-4mm}
\end{center}
\end{table*}

\section{Concerning Learnability}
It is noticed in Section~\ref{sec: schedule} that the role of $\lambda$ is two-fold. On the one hand, it controls the robustness of the model to mislabeling in the dataset. On the other hand, it controls the learnability of the training set. In fact, experiments reveal that the phenomenon is quite dramatic, in the sense that a phase transition-like behavior exists when different $\lambda$ is used.

\begin{figure}

         \centering
         \includegraphics[width=0.45\textwidth]{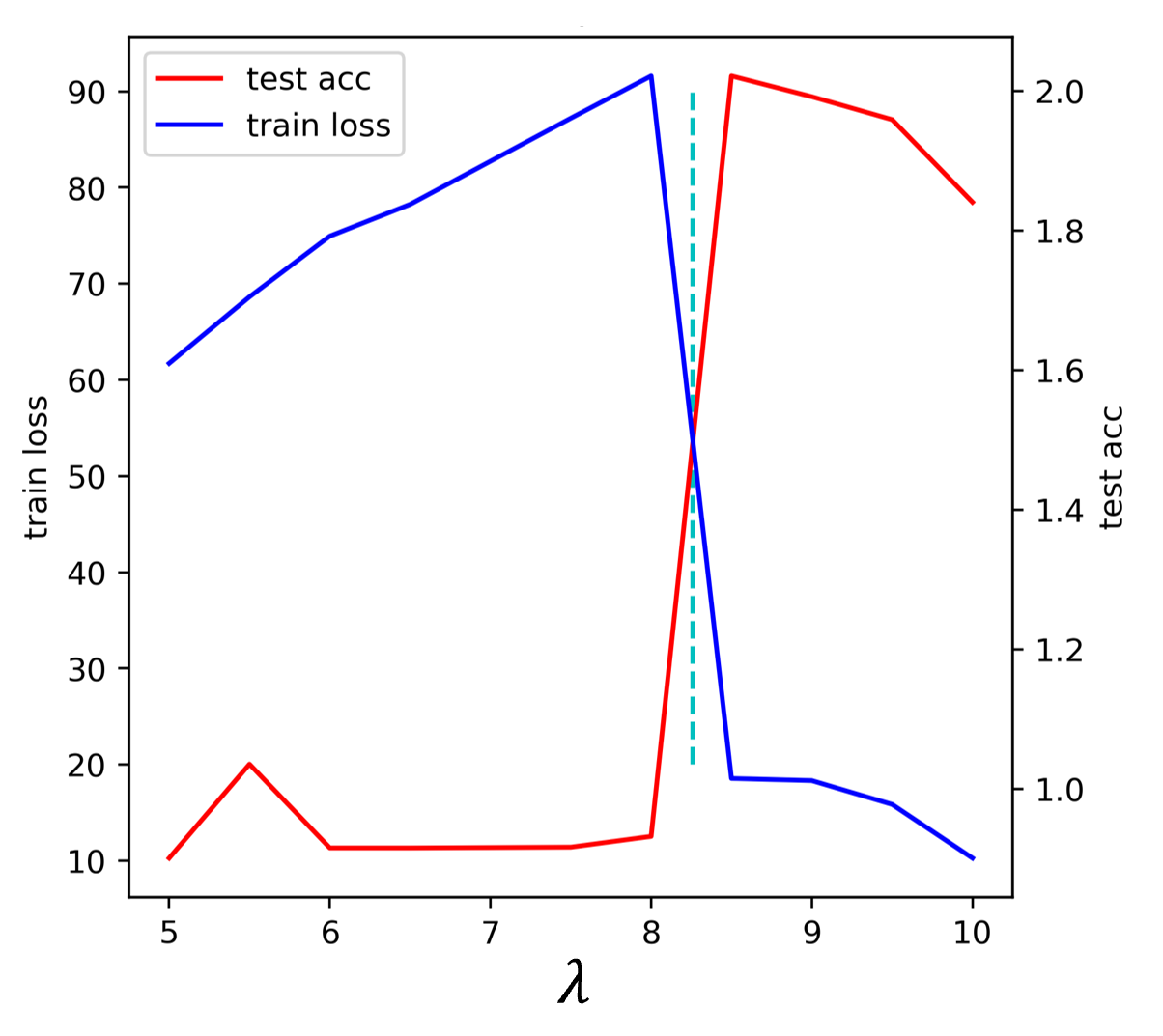}
         \vspace{-7mm}
     \caption{Critical behavior of the gambler's loss. Data showing that the learning almost do not happen at all for $\lambda<\lambda_{\mathrm{crit}}$, while above $\lambda_{\mathrm{crit}}$ the behavior is qualitatively similar. The optimal $\lambda$ can be tuned for using this phenomenon; however, more often one does not need to tune for o.\vspace{-4mm}}
     \label{fig: critical}
\end{figure}

In particular, we note that there is a ``good" range for hyperparameter $\lambda$ and a bad range. The good range is between a critical value $\lambda_\mathrm{crit}$ and $M$ (non-inclusive) and the bad range is smaller than $\lambda_\mathrm{crit}$. See Figure~\ref{fig: critical} for an example on MNIST with $r=0.5$. We see that, in the good range, reducing $\lambda$ improves robustness, resultin in performance improvement by more than $10\%$ accuracy; however, reducing $\lambda$ below $\lambda_{\mathrm{crit}}$ makes learning impossible. In this example, we experimentally pin-down $\lambda_{\mathrm{crit}}$ to lie in $(8.20,\ 8.21)$.

\end{document}